\documentclass[conference]{IEEEtran}
\IEEEoverridecommandlockouts
\usepackage{cite}
\usepackage{amsmath,amssymb,amsfonts}
\usepackage{algorithmic}
\usepackage{graphicx}
\usepackage{textcomp}
\usepackage{xcolor}
\usepackage{hyperref}
\usepackage{blindtext}
\usepackage{diagbox}

\def\BibTeX{{\rm B\kern-.05em{\sc i\kern-.025em b}\kern-.08em
    T\kern-.1667em\lower.7ex\hbox{E}\kern-.125emX}}
\begin{document}

\title{Reinforcement Learning for Predicting Traffic Accidents}
\makeatletter
\newcommand{\linebreakand}{%
  \end{@IEEEauthorhalign}
  \hfill\mbox{}\par
  \mbox{}\hfill\begin{@IEEEauthorhalign}
}
\makeatother

\author{
\IEEEauthorblockN{Injoon Cho}
\IEEEauthorblockA{\textit{Cho Chun Shik Graduate School of Mobility} \\
\textit{Korea Advanced Institute of Science and Technology}\\
Daejeon 34141, Republic of Korea \\
oprij12@kaist.ac.kr}
\and
\IEEEauthorblockN{Praveen Kumar Rajendran}
\IEEEauthorblockA{\textit{Division of Future Vehicle} \\
\textit{Korea Advanced Institute of Science and Technology}\\
Daejeon 34141, Republic of Korea \\
praveenkumar@kaist.ac.kr}
\linebreakand
\IEEEauthorblockN{Taeyoung Kim}
\IEEEauthorblockA{\textit{Cho Chun Shik Graduate School of Mobility} \\
\textit{Korea Advanced Institute of Science and Technology}\\
Daejeon 34141, Republic of Korea \\
ngng9957@kaist.ac.kr}
\and
\IEEEauthorblockN{Dongsoo Har}
\IEEEauthorblockA{\textit{Cho Chun Shik Graduate School of Mobility} \\
\textit{Korea Advanced Institute of Science and Technology}\\
Daejeon 34141, Republic of Korea \\
dshar@kaist.ac.kr}
}

\maketitle

\begin{abstract}
As the demand for autonomous driving increases, it is paramount to ensure safety. Early accident prediction using deep learning methods for driving safety has recently gained much attention. In this task, early accident prediction and a point prediction of where the drivers should look are determined, with the dashcam video as input. We propose to exploit the double actors and regularized critics (DARC) method, for the first time, on this accident forecasting platform. We derive inspiration from DARC since it is currently a state-of-the-art reinforcement learning (RL) model on continuous action space suitable for accident anticipation. Results show that by utilizing DARC, we can make predictions 5\% earlier on average while improving in multiple metrics of precision compared to existing methods. The results imply that using our RL-based problem formulation could significantly increase the safety of autonomous driving. 
\end{abstract}


\section{Introduction}
The automotive industry is increasingly paying significant attention to autonomous vehicle technology. It is expected that vehicles with level 4 or higher autonomous driving \cite{sae2014automated} will be commercialized and take over most of the roads very soon. As the interest in autonomous vehicles grows, safety problems of autonomous vehicles are earning more spotlight. Although various autonomous applications are in place for commercialized vehicles, there still are problems related to driver safety that must be addressed \cite{wang2020safety}. Sensors such as measuring the distance to an object, such as sonar, LiDAR, or radar, and algorithms such as optimal path planning algorithms, exist solely for safety reasons. 

Among technologies for safe autonomous driving, accident prediction and detection have a higher priority due to their potential for saving lives. Numerous accidents come from late reactions of drivers in unexpected situations \cite{drozdziel2020drivers}. To avoid accidents on the road, early anticipation of traffic accidents is essential. Anticipating accidents from inside the car itself can aid the drivers in avoiding accidents and further help the autonomous agent to make optimal and safe decisions in risky situations. For studies on anticipating accidents, dashcam video datasets are used broadly. Since dashcams are commonly used in automobiles these days, gathering and employing data are straightforward \cite{chan2016anticipating}. The anticipation results from dashcam data can aid the driver to be more aware of the environment by Advanced Driver Assistance Systems (ADAS) or make higher level accident prevention possible for autonomous vehicles. Extending the time between the actual accident and the prediction by even a few seconds will prevent a considerable amount of traffic accidents.

\begin{figure}[t]
    \centerline{\includegraphics[width=0.50\textwidth]{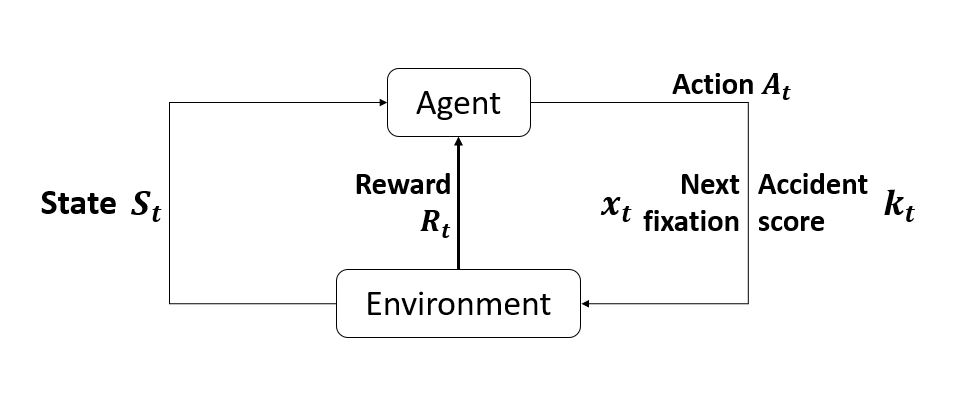}}
    \caption{Markov Decision Process structure in this work differs from the renowned MDP structure. The action space works with two policies.} 
    \label{frameworks}
\end{figure}

Accident anticipation can be divided into two tasks, namely, early anticipation and accurate anticipation. Previous studies have been attempting to predict traffic accidents as early as possible while keeping the accuracy of the prediction as high as possible \cite{chan2016anticipating, bao2020ustring}. However, there exists a trade-off between the two tasks. For the prediction to be more accurate, it is better to make the prediction as late as possible. To make an early decision, the anticipation accuracy drops eventually.

Consequently, existing studies on accident anticipation adopt deep learning to compensate for the two tasks and obtain the best result possible in autonomous driving \cite{chan2016anticipating, bao2020ustring}. Predicting whether the accident will happen or not is a classification problem. Since supervised learning is widely used for classification problems, the majority of the work that has been done in this field has adopted supervised learning. Furthermore, a recent study on traffic accident anticipation moved on to considering fixation prediction \cite{bao2021drive}, which is the point where the drivers will look at the next time-step. This makes two tasks for accident anticipation: accident anticipation and fixation prediction. The study \cite{bao2021drive} used the Markov decision process (MDP) to formulate the tasks in one scenario and developed it into deep reinforcement learning (DRL) problem, which as a result, outperformed almost every existing state-of-the-art accident anticipation model. 

This paper adopts an enhanced version of the DRL algorithm to anticipate accidents earlier and more accurately than state-of-the-art models. The double actor regularized critics (DARC) \cite{lyu2022darc}, proposed in 2021, showed outstanding performance in continuous action spaces. DARC has a twin delayed deterministic (TD3) form with an additional actor network, which encourages more exploration of the agent and helps to control the overestimation bias simultaneously. Due to the additional actor network, double critic in TD3 can be obsolete. DARC solves this problem by regularization of critics. This regularization enables value errors, policy execution errors, and critic deviance errors, all in control to outperform state-of-the-art DRL algorithms. The DRL model structure, including the observation state and the reward, is based on the DRIVE model \cite{bao2021drive}.

The proposed approach to the anticipation of an accident using dashcam videos is distinctive since most of the accident anticipation models follow supervised learning (SL) except the DRIVE model. This paper enhances the accident anticipation model based on DRL a step forward. The results show that with this model applied to ADAS or any other autonomous vehicle applications, it would significantly reduce traffic accidents. The main contributions of this study are as follows:
\begin{itemize}
\item The state-of-the-art deep reinforcement learning algorithm is utilized in the accident anticipation platform and outperforms existing methods of accident anticipation.
\item We examine the improvement of the existing actor-critic algorithms with additional actor network incorporation for the accident anticipation task. 
\end{itemize}

\section{Related Works}

\subsection{Reinforcement Learning}
Reinforcement learning is a part of the machine learning paradigm and is based on trial and error to make decisions. Based on the environment and state, the agent chooses the action to maximize the accumulated reward. Many reinforcement learning algorithms exist to optimize the policy that decides which actions to choose at every step, and policy gradient methods are among them. Policy gradient is an algorithm that tries to boost the probabilities of the actions that can bring higher rewards overall. This work execution follows continuous action spaces, so we focus on the policy gradient algorithms implemented on continuous action spaces. 

Deep deterministic policy gradient (DDPG) is an algorithm that learns policy simultaneously with Q-function \cite{lillicrap2015continuous}. With the actor-critic method's structure, DDPG outruns the performance of DQN. However, there can be an overestimation bias in DDPG, which is addressed by adding additional criticism to the algorithm. This makes twin-delayed DDPG (TD3) significantly outperform DDPG \cite{fujimoto2018td3}. In the same year, when TD3 was presented, soft actor-critic (SAC) was also proposed. Soft actor-critic optimizes stochastic policy and adopts an entropy term to enhance the agent's experience, producing high-performance \cite{haarnoja2018sac}. Until recently, TD3 and SAC were considered the most advanced algorithms for model-free deep reinforcement learning. It also acted as the basis for different algorithms \cite{review}.

Studies on reinforcement learning also discovered a method using double actors. In \cite{lyu2022darc}, dual actors and regularized critics are proposed, showing its improvement in the MuJoCo platform. By using double actors, the critics can choose from two policies, $\pi_{\phi_1'}$ and $\pi_{\phi_2'}$. Selecting the policy that makes the minimum value estimation mitigates the overestimation problem in DDPG. When the critics are doubled as well as actors, we can choose the maximal one for the final value estimation, which will address the underestimation problem in TD3. Updating the value function in this way is described in \eqref{eq:value}. However, two independent critics can bring uncertainty in value estimation. Therefore, DARC defines three errors: value error $\epsilon_k$, policy execution error $\epsilon_\pi$, and critic deviance error $\epsilon_d$, and controls them by regularizing critics. Value estimation and critic regularization are shown in \eqref{eq:errors}, where $\nu$ is the soft update rate and $\gamma$ is the discount factor.

{\scriptsize
\begin{gather}
\label{eq:value}
    \hat{V}(s^\prime) \leftarrow \max \left\{ \min_{i=1,2}\left(Q_{\theta_i^\prime}(s^\prime,\pi_{\phi_1^\prime}(s^\prime))\right), \min_{j=1,2}\left(Q_{\theta_j^\prime}(s^\prime,\pi_{\phi_2^\prime}(s^\prime))\right) \right\},
\end{gather}
}

{\scriptsize
\begin{gather}
\label{eq:errors}
    \begin{split}
        \| V_t(s) - V^*(s)\|_\infty \le \gamma^t\|V_0(s) - V^*(s)\|_\infty + \\
        \sum_{k=0}^t \gamma^k \epsilon_k +  \dfrac{\nu}{1-\gamma}(2\epsilon_d+\epsilon_\pi).
    \end{split}
\end{gather}
}
In MuJoCo environments, TD3 \cite{fujimoto2018td3} and SAC \cite{haarnoja2018sac} show 245\% and 191\% average improvements, based on deep deterministic policy gradient (DDPG), respectively. DARC offers 331\% of average improvement in the same condition. The double actors significantly boost the exploration of the agent, making the model converge quickly and obtain the highest return.

Compared to supervised learning, reinforcement learning can result in more advanced results than human-level performance. For this reason, reinforcement learning study takes different paths. Some research is on the algorithm's structure, such as \cite{seo2019rewards} on rewards and \cite{vecchietti2020sampling} on replay buffers. \cite{vecchietti2020batch} investigated the model, which has multiple goals, like in this paper. DRL can have applications in various fields, such as path planning \cite{moon2022path}, robot soccer games \cite{kim2021two}, or autonomous driving \cite{kim2021reinforcement}. 

\subsection{Traffic Accident Anticipation}

Starting from \cite{chan2016anticipating}, which used DSA-RNN for anticipating accidents in dashcam videos, studies on predicting accidents using dashcam videos are investigated. The study showed better results for more accurate prediction when dynamic spatial and temporal attention was used \cite{monjurul2021dsta}. Accurate prediction and early anticipation are challenging to achieve as a joint task, as shown by the studies above. Concerning the trade-off, \cite{bao2021drive} formed MDP based DRL model, DRIVE, which outperformed every model with supervised learning.

\section{Methodology}

The overview of the reinforcement learning structure is as follows. \figureautorefname{2} shows the outline of the model proposed in this study. Since the task of this model is to predict the accident as early as possible while keeping the prediction accuracy high, problem settings are based on existing research \cite{bao2021drive}. With dashcam video data as input, the saliency model processes the input to form the environment for the machine learning agent. The main structure of the machine learning model is based on MDP, while the environment of the MDP consists of attention made from foveal vision. Shaping up the environment as done in \cite{bao2021drive} will give us attention to more risky regions in the frame, which imitates the behaviour of human perception. Models following the MDP usually assume that a reward at each time step follows an action. However, in this model, based on the MDP structure, there are two actions, $k_t$, and $x_t$. Specifically, $k_t$ and $x_t$ represent the prediction of accident and prediction of fixation, respectively. Since the agent selects two actions every step, there should be two corresponding rewards which are denoted by $r_A$ and $r_f$. Conclusively, this model learns to predict the probability of traffic accidents while recognizing the surrounding environment using the human-imitating attention model. 

\begin{figure}[t]
    \centerline{\includegraphics[width=0.50\textwidth]{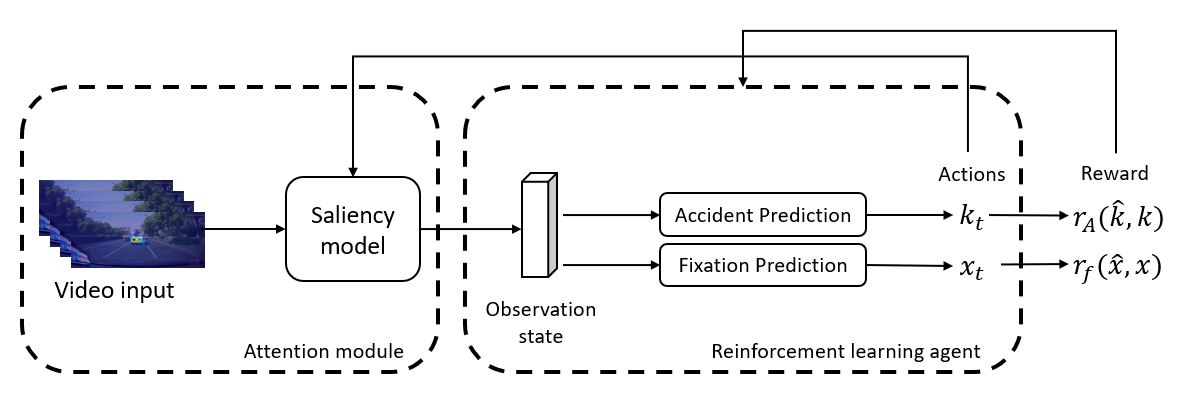}}
    \caption{The attention model gets the video dataset as input and processes it through an already-trained saliency model. The result works as the observation state for the reinforcement learning model. Having two tasks, the reinforcement learning model trains with two policies alongside each other.} 
    \label{frameworks}
\end{figure}

For the attention part of the model, we mostly follow the DRIVE model structure. For traffic accident prediction, examining and analyzing the entire portion of each video frame can take time and a lot of computation. We can use a method that imitates the human perception mechanism to avoid inefficiency, as discussed in \cite{bao2021drive}. Visual perception of human beings can be explained in two ways: top-down attention and bottom-up attention. Initially, top-down attention allocates its attention to certain features first and then spreads to the rest of the frame. On the other hand, bottom-up attention starts from the whole frame and narrows down its attention to the essential features. By combining the two ways of visual attention, we can obtain a model processing similar to how the human visual perception mechanism works. A convolutional neural network (CNN) based saliency module processes input data for bottom-up attention. For top-down attention, input data goes through the foveal vision module before going into the saliency module. Finally, both attentions are normalized and combined with the empirical ratio to complete our reinforcement learning model's observation environment. 

Our model's action space correlates with the accident score and the subsequent fixation. The concatenation of the accident score policy and the fixation policy form the action space. During training, the two policy networks and two fully connected layers with ReLU activation determines the action. Positioning the long short-term memory (LSTM) layer after the last layer of fully connected layers, we can expect the model to acquire the temporal dependency of connected actions. As mentioned, the agent should solve the multi-task problem to simultaneously perform accident anticipation and fixation prediction. The rewards of the agent are adopted from~\cite{bao2021drive}. For the reward functions of accident score, we use the XNOR gate so that it assigns a reward of 1 to the accurate predictions, whereas 0 to the false predictions. 

\begin{figure}[t]
    \centerline{\includegraphics[width=0.50\textwidth]{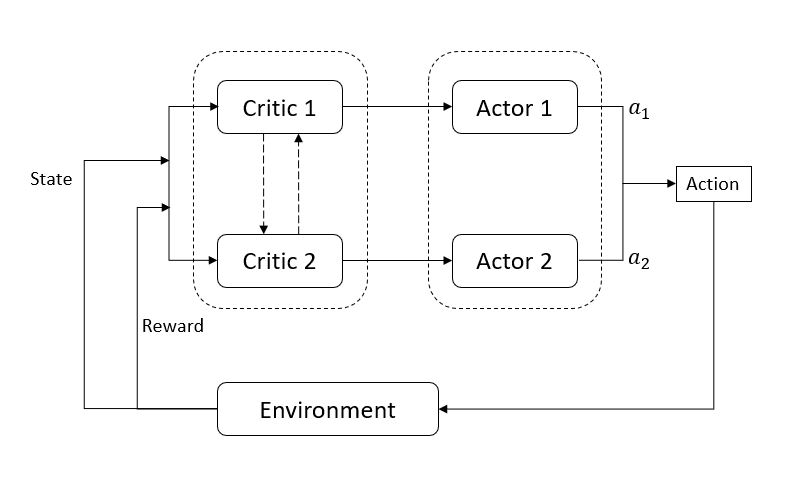}}
    \caption{The structure is similar to that of TD3 but has an additional actor network, and the critics are linked with each other in order to regularize.} 
    \label{frameworks}
\end{figure}

\begin{gather}
\label{eq:r_A}
    r_A^t = w_t \cdot \text{XNOR} \left[\mathbb{I}[a^t > a_0], y\right],
\end{gather}

With the XNOR gate, we use the weighting factor $w_t$ and an indicator function $\mathbb{I}$ to form the reward function for the accident score. The weighting factor is designed to decay exponentially and becomes 0 at the point where the accident occurs. 

\begin{gather}
    w_t = \frac{1}{e^{t_a}-1} \left(e^{\max(0, t_a-t)} - 1\right),
\label{earliness}
\end{gather}

This improves the earliness of the prediction. A different design is in place for fixation reward \eqref{eqn:r_f}. The reward function uses 2-D coordinates of the predicted fixation point and ground truth fixation point, $\hat{p_t}$ and $p_t$. The agent is rewarded by how close the ground truth and predicted fixation point is. $\eta$ is a coefficient empirically obtained to maintain a balance between fixation and accident rewards.

\begin{gather}
    r_F^t = \mathbb{I}\left[t>t_a\right] \exp{\left(-\frac{||\hat{p}^t - p^t||^2}{\eta}\right)},
\label{eqn:r_f}
\end{gather}

Our method trains the model with double actors and regularized critics (DARC) \cite{lyu2022darc}, a recent state-of-the-art deep reinforcement learning algorithm. Figure 3 shows the structure of the algorithm. By using double actors, the critics can choose from two policies. In our work, this makes two policies for accident score and another two policies for fixation prediction, which makes four policies. As in Section B, we follow the regularization steps for the critics.

\section{Results}

\subsection{Implementation Details}

The model proposed in this paper is built upon ubuntu 20.04, with AMD Ryzen 7 1800X Eight-Core Processor as CPU and GeForce GTX 1080 Ti as GPU device. For the saliency module, we adopted VGG-16-based MLNet \cite{cornia2016deep}, which was also used in \cite{bao2021drive}. Source code was based on the DRIVE model \footnote[1]{\url{https://github.com/Cogito2012/DRIVE}}\cite{bao2021drive}, TD3 model \footnote[2]{\url{https://github.com/sfujim/TD3}}\cite{fujimoto2018td3}, and DARC model \footnote[3]{\url{https://github.com/dmksjfl/DARC}}\cite{lyu2022darc}. This work used the Adam optimizer for every gradient descent, and we trained every algorithm for 30 epochs. 

This work uses the DADA-2000 dataset \cite{dada-2000} to evaluate the model's performance. DADA is an abbreviation for Driver Attention in Driving Accident scenarios, which has data from dashcam videos containing driving accidents. 

\subsection{Results and Discussion}

For the evaluation of the model, we adopt several metrics existing in machine learning classification and the computer vision field. For classification models, receiver operating characteristics and Precision-Recall (PR) curve is primarily used to evaluate the performance. ROC is a probability curve drawn at various threshold values, which by setting True Positive Rate (TPR) as the y-axis and False Positive Rate (FPR) as the x-axis, separates signal from noise. One can use the area under the curve (AUC) to evaluate the correctness of the accident anticipation. The value of AUC ranges from 0 to 1, and the closer to 1 the AUC value is, the better the performance of the classification model is. Having precision and recall (TPR) as axes, the PR curve is also used for assessing the classification ability of the algorithm by taking the area below the curve. This area is called average precision (AP). Like AUC, AP is also desired to be close to 1. Aside from the model's correctness, the earliness of prediction is evaluated by mean time-to-accident (mTTA). The time of the accident is calculated by subtracting the prediction time from the accident time. Time-to-accident is set to zero when the accident prediction happens during or after the accident.

% \usepackage{graphicx}
\begin{table}[ht]
\centering
\resizebox{\columnwidth}{!}{%
{\footnotesize
\begin{tabular}{|c|c|c|c|c|} \hline
  \diagbox{Metrics}{Methods}   & SAC    & DDPG & TD3    & DARC            \\ \hline
mTTA & 3.5455 &   3.4046   & 2.4612 & \textbf{3.7238} \\ \hline
AUC & 0.62744 &  0.61428    & 0.57576 & \textbf{0.70793} \\
AP & 0.847 &    0.8544    & 0.81801 & \textbf{0.8621}  \\
recall & 0.9043 & 0.8764 & 0.8625 & \textbf{0.9284} \\
fixationMSE & 134.0964 & 136.0483  & 301.5777 & \textbf{133.0469}\\

 \hline
\end{tabular}%
}
}
\vspace{0.1cm}
\caption{Comparison of earliness and correctness of algorithms.}
\label{tab:my-table}
\end{table}

% \diagbox{Input}{Output}
\begin{center}
\begin{figure}[t]
    \centerline{\includegraphics[width=0.50\textwidth]{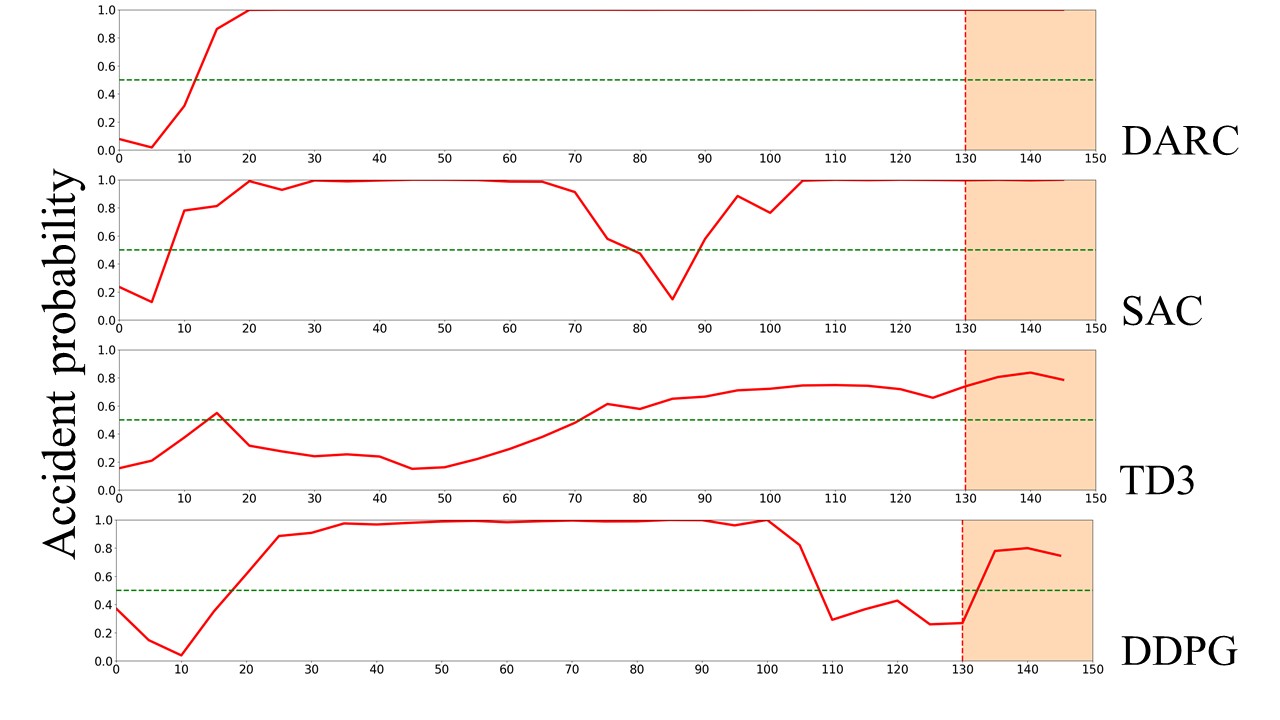}}
    \caption{The predicted accident probability of each algorithm is depicted for every frame in this graph. DARC model not only shows earlier prediction than other algorithms but also shows the robustness of the prediction.} 
    \label{frameworks}
\end{figure}\end{center}

Previous accident anticipation work done with reinforcement learning algorithm \cite{bao2021drive} set the baseline as existing supervised learning models \cite{bao2020ustring, chan2016anticipating} as well as the accident anticipation loss function AdaLEA \cite{suzuki2018anticipating}. In this study, the results from the soft actor-critic, used in the DRIVE model, are set as the baseline. The results are shown in \tableautorefname{1}. Our proposed model utilizing DARC was able to exceed the accident anticipation earliness of other famous algorithms and further exceed the DRIVE model's performance. The details of the accident prediction are depicted in \figureautorefname{4}. Prediction from DARC shows the most earliness as well as robustness.

The research on the reaction time of drivers on the road \cite{drozdziel2020drivers} gives us the minimum length of time for the drivers to be prepared for the accident. The perception time, primary psychological reaction time, and foot transfer time of the driver sums up to be average of 2 seconds. This implies that drivers need at least 2 seconds to react to unexpected situations. Because of this, going well beyond maintaining the correctness of the model was very important. Since mTTA was already in the range of safe-reaction time, DRIVE model can already achieve more than 2 seconds of mTTA. This implies that we should focus more on the correctness of the results. 

While achieving earlier prediction on average, our model managed to maintain or obtain comparatively higher accuracy than existing models. As can be seen from \tableautorefname{1}, AUC and AP values show that the DARC algorithm has better classification accuracy than other algorithms. Furthermore, we can see that the recall value of DARC shows superior results over different algorithms. Since recall is calculated by TP / (TP + FN), a low recall value indicates that there were accidents that occurred without being anticipated. Therefore, recall value is one of the most important values to be considered for safety. 

\begin{figure}[t]
    \centerline{\includegraphics[width=0.50\textwidth]{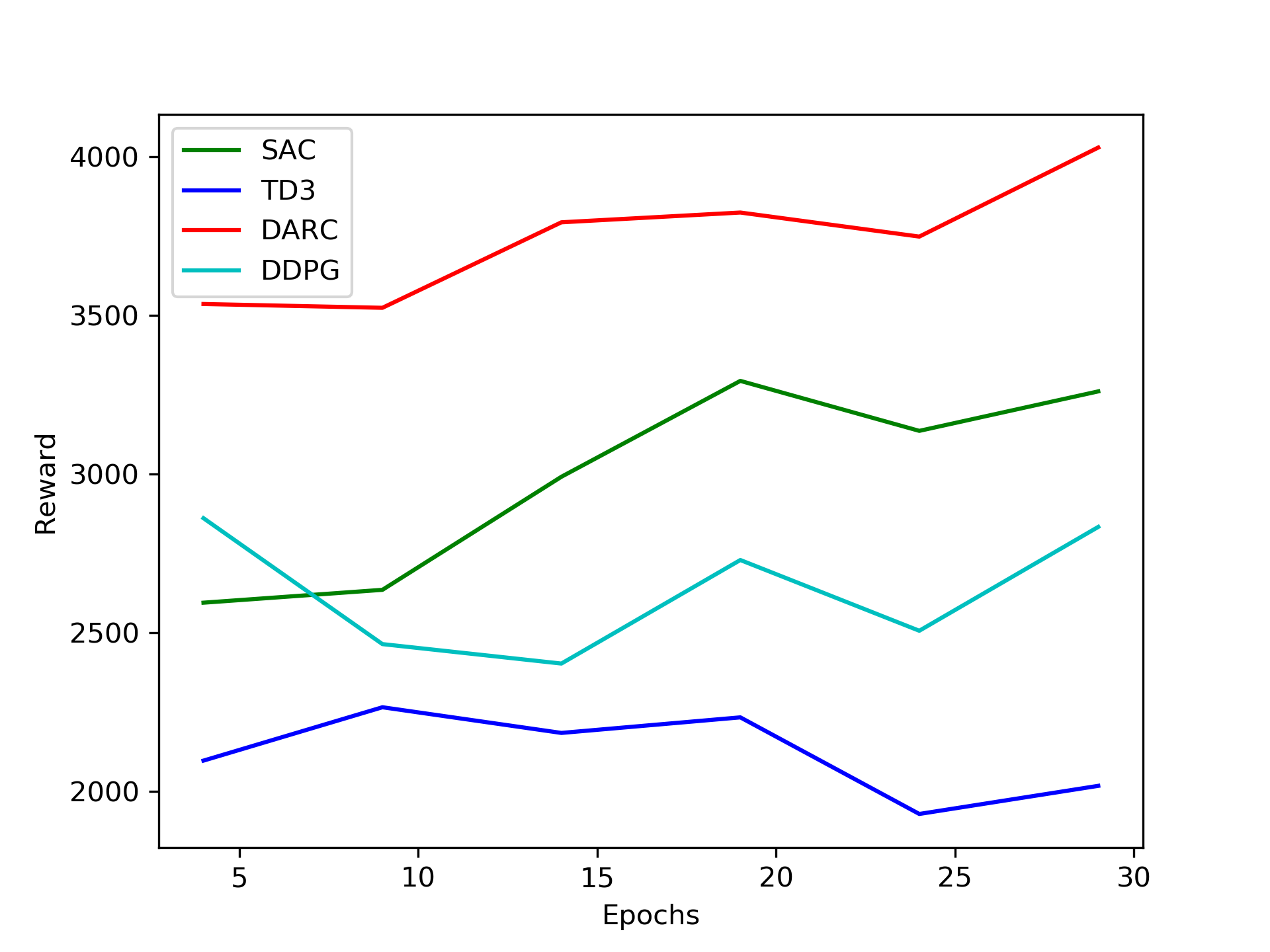}}
    \caption{The graph shows that the DARC algorithm easily outruns the performance of other algorithms. In the test results, the rewards were recorded every five epochs.} 
    \label{frameworks}
\end{figure}

For the comparison of reinforcement learning performances in the accident anticipation platform, it is clearly visible in \figureautorefname{5} that the DARC algorithm outperforms existing state-of-the-art algorithms. This result suggests that DARC can also be generalized to the real world and safety critical applications like accident anticipation. 

\section{Conclusion}

This paper proposes a reinforcement learning model for early and accurate accident anticipation in autonomous vehicles. Our model utilizes a new actor-critic method DARC, and the experiments showed that the model exceeds the performances of existing models. By using multiple actors, the model manages to outperform the exploration of the entropy regularization-based SAC algorithm, and by updating the critics in a novel method, the model tends to minimize both overestimation and underestimation bias more than the TD3 algorithm. Moreover, results in anticipating accidents show the effect of increasing actors and critics in the modified actor-critic method using the task as a reinforcement learning platform. In addition to ensuring the safety of autonomous vehicles, this research not only shows the adaptability of reinforcement learning algorithms but also suggests the possibilities of usage in other fields that have only been using supervised learning.

\section*{Acknowledgment}

This work was supported by the Institute for Information communications Technology Promotion (IITP) grant funded by the Korean government (MSIT) (No.2020-0-00440, Development of Artificial Intelligence Technology that continuously improves itself as the situation changes in the real world).

\bibliographystyle{plain}
\bibliography{references}

\end{document}